\documentclass{article}
\usepackage{ijcai24}
\usepackage{marvosym}
\usepackage{times}
\usepackage{soul}
\usepackage{url}
\usepackage[hidelinks]{hyperref}
\usepackage[utf8]{inputenc}
\usepackage[small]{caption}
\usepackage{graphicx}
\usepackage{amsmath}
\usepackage{amsthm}
\usepackage{booktabs}
\usepackage{algorithm}
\usepackage{algorithmic}
\usepackage[switch]{lineno}
\usepackage{algorithm}
\usepackage{algorithmic}
\usepackage{graphicx} 
\usepackage{amsmath}  
\usepackage{graphicx} 
\usepackage{multirow} 
\usepackage[table]{xcolor} 
\usepackage{amssymb} 
\usepackage{hyperref}

\title{Exploring Color Invariance through Image-Level Ensemble Learning}

\author {
	Yunpeng Gong$^1$
	\and
	Jiaquan Li$^1$\and
	Lifei Chen$^2$\and
	Min Jiang$^{1}${\Letter}
	\affiliations
	$^1$School of Informatics, Xiamen University, China\\
	$^2$College of Computer and Cyber Security, Fujian Normal University, China
	\emails
	\ fmonkey625@gmail.com,
	\ 31520211154061@stu.xmu.edu.cn,
	\ clfei@fjnu.edu.cn,
	\ minjiang@xmu.edu.cn
}
\begin{document}
	
\maketitle
	
\begin{abstract}
In the field of computer vision, the persistent presence of color bias, resulting from fluctuations in real-world lighting and camera conditions, presents a substantial challenge to the robustness of models. This issue is particularly pronounced in complex wide-area surveillance scenarios, such as person re-identification and industrial dust segmentation, where models often experience a decline in performance due to overfitting on color information during training, given the presence of environmental variations. Consequently, there is a need to effectively adapt models to cope with the complexities of camera conditions. To address this challenge, this study introduces a learning strategy named Random Color Erasing, which draws inspiration from ensemble learning. This strategy selectively erases partial or complete color information in the training data without disrupting the original image structure, thereby achieving a balanced weighting of color features and other features within the neural network. This approach mitigates the risk of overfitting and enhances the model's ability to handle color variation, thereby improving its overall robustness. The approach we propose serves as an ensemble learning strategy, characterized by robust interpretability. A comprehensive analysis of this methodology is presented in this paper. Across various tasks such as person re-identification and semantic segmentation, our approach consistently improves strong baseline methods. Notably, in comparison to existing methods that prioritize color robustness, our strategy significantly enhances performance in cross-domain scenarios. The code available at \url{https://github.com/layumi/Person\_reID\_baseline\_pytorch/blob/master/random\_erasing.py} or \url{https://github.com/finger-monkey/Data-Augmentation}.
\end{abstract}

\section{Introduction}
With the rapid advancement of computer vision and deep learning techniques, wide-area surveillance \cite{yuan2023lightweight,Gong_2022_CVPR,gong2024} systems are crucial in various applications, such as public safety and industrial monitoring. Wide-area surveillance involves the monitoring and analysis of large-scale environments through diverse camera setups, providing critical information for decision-making and security. One of the primary challenges faced by deep learning models in such scenarios is the presence of color deviation arising from variation camera environments.

In wide-area surveillance scenarios, models are susceptible to color domain deviations due to diverse camera settings, complex lighting, and extensive scene coverage. These deviations are exacerbated by variations in lighting and environmental conditions, impacting color representation \cite{Camera-style}. This problems can lead to model instability and compromise performance in tasks such as person re-identification and industrial dust segmentation.

\begin{figure}[]
	\setlength{\belowcaptionskip}{-0.4cm}   
	\centering
	\includegraphics[width=1\linewidth]{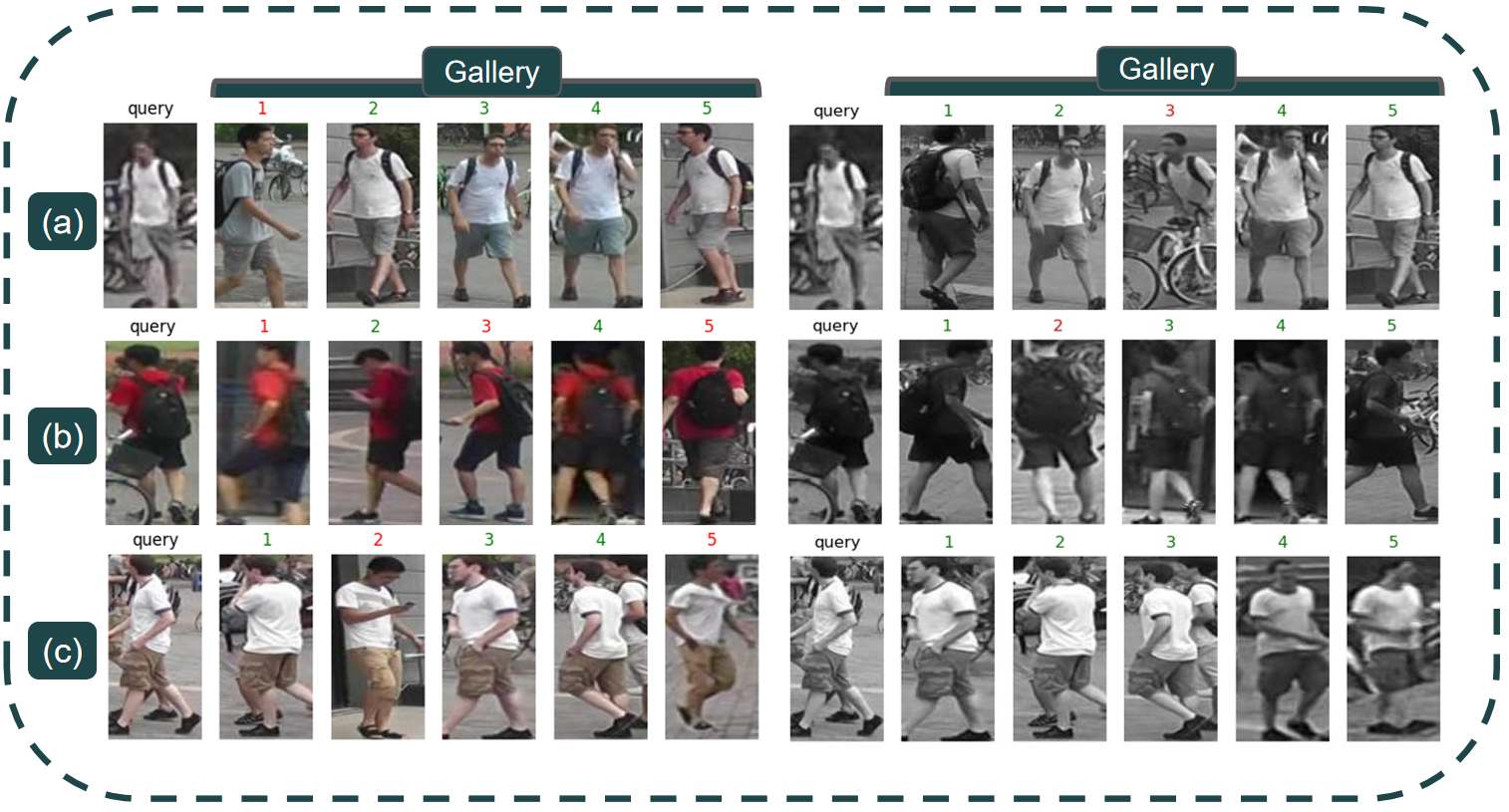}
	\caption{The retrieval results of the model trained with visible (RGB) image and the model trained with grayscale image on the Market1501 dataset. The numbers on the images indicate the rank of similarity in the retrieval results, the red and green numbers denote the wrong and correct results, respectively.}
	\label{fig1}
\end{figure}

\begin{figure*}[t]
	\setlength{\abovecaptionskip}{0.1cm}
	\setlength{\belowcaptionskip}{-0.4cm}   
	\centering
	\includegraphics[width=1\linewidth]{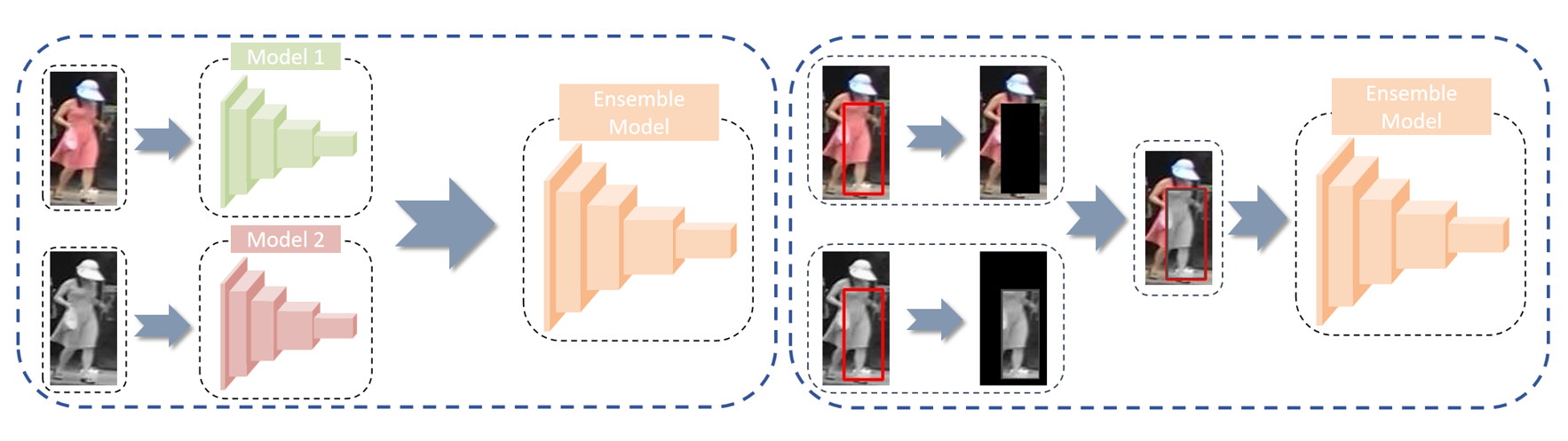}
	\caption{The figure illustrates two approaches for learning color invariance. The left image demonstrates the integrated model trained with samples containing and lacking color information separately. On the right, our proposed image-level ensemble learning strategy, Random Color Erasing, achieves color invariance by applying random local homogeneous grayscale transformations to integrate samples with and without color information simultaneously, thereby achieving equivalent effects to ensemble models}
	\label{LGT-LST}
\end{figure*}

\subsubsection{Person Re-identification}
Person re-identification (ReID) is a challenging computer vision task focused on recognizing individuals across camera views \cite{zhou2023adaptive}. The core challenge of person re-identification lies in the variations introduced by pose changes, camera viewpoints, occlusions, and lighting conditions etc., resulting in significant intra-class differences \cite{zheng2019joint}. This variation often leads to substantial changes in the appearance of the same pedestrian image, causing substantial metric distance differences. Consequently, ReID models may exhibit an excessive reliance on color information, which is often the most salient and easily distinguishable feature, as emphasized by Gong \cite{Gong_2022_CVPR}. This reliance stems from the challenge of feature matching.

It is no doubt that color features are important discriminative features, but it will restrict the model to make correct predictions in some cases. Fig.~\ref{fig1} demonstrates that the color deviation between query and gallery images impacts retrieval outcomes, and, interestingly, some samples exhibit improved retrieval results when color information is ignored.

\subsubsection{Semantic Segmentation}
In specific industrial production settings such as construction sites, mining fields, and manufacturing factories, dust (or smoke) represents a prevalent issue, significantly impacting construction progress and the health of workers.  

Industrial dust (or smoke) segmentation \cite{yuan2023lightweight} poses another highly challenging task in the realm of wide-area surveillance scenarios. Smoke or dust exhibits dynamic morphological variations as it flows and diffuses in the air due to its minute particles and gaseous nature. Its shape and distribution are influenced by surrounding environments, airflow, and temperature, leading to irregular and unstable characteristics in spatial representation. These factors also make the model overly reliant on color information, resulting in insufficient robustness of the model to color deviations. Due to the aforementioned challenges, boundary regions often present difficulties for segmentation models, as they tend to exhibit prediction errors, leading to suboptimal performance in practical applications. 

To tackle the aforementioned challenge, this paper introduces an innovative strategy known as Random Color Erasing (RCE). RCE is devised to restore the balance between color features and other crucial discriminative features within a neural network. This strategy can manifest in various manifestations. RCE involves the random selection of an area within an RGB image and the replacement of its pixels with the corresponding area from a grayscale image (or vice versa, selecting an area within a grayscale image and substituting its pixels with the corresponding pixels from the equivalent region in an RGB image). This process generates new training samples with varying degrees of homogeneously mixed domains, contributing to model robustness during training. In comparison to existing techniques based on Generative Adversarial Networks (GANs), the proposed approach demonstrates superior efficiency and effectiveness. Notably, it avoids introducing additional noise while conserving substantial computational resources.

In addition, this paper analyzes the relationship between proposed RCE and the generalization ability of neural networks from the perspective of classification, and reveals the intrinsic reasons that networks trainning with RCE may outperform ordinary networks. Experiments show that the proposed method increases the robustness of the model to color deviation, and bridges the domain gap between different datasets. 

The main contributions of this paper are summarized as follows:

$\bullet$ This paper introduces the Random Color Erasing (RCE) which enhances color-invariant learning in varied visual scenarios. This strategy effectively mitigates overfitting, thereby bolstering the model's ability for improved generalization.

$\bullet$ This paper conducts an analysis that the network trained with RCE may be better than the ordinary network from the perspective of classification.

$\bullet$ The extensive experiments on two distinct visual tasks and five baseline models with diverse architectures demonstrate the effectiveness of the proposed ensemble learning strategy. Particularly, our method exhibits significant potential surpassing traditional approaches in cross-domain testing. This approach provides a novel perspective for subsequent research.
\section{Related Works}
Over the years, ReID has been devoted to seeking effective solutions to enhance the model's robustness to color variations. To address this, numerous methods have been introduced and proposed. They can be broadly categorized into classical approaches and GAN-based methods.

\subsection{Classical Methods}
In solving the problem of color deviation, the early work \cite{li2014deepreid} used the filter and the maximum grouping layer to learn the illumination transformation, divided the pedestrian image into more small pieces to calculate the similarity, and uniformly handled the problems of misalignment, occlusion and illumination variation under the deep neural network. Liao et al. \cite{liao2015person} performed pre-processing before feature extraction and used multiscale Retinex algorithm to enhance the color information of light shaded regions to improve the color changes caused by lighting condition changes.

Many data augmentation methods have been proposed, such as random cropping, flipping, which are widely used in computer vision. Different methods address distinct issues. CutMix \cite{yun2019cutmix} replaces one patch
of an image with a patch from another image, it helps improve the robustness and generalization of deep learning models by encouraging them to learn from mixed and diverse training samples. Autoaugment \cite{47890} is a automated data augmentation strategy, which incorporates a series of data augmentation such as image shear transformations, color adjustments, brightness/contrast adjustments, etc..  Co-Mixup \cite{kim2021comixup} combines CutMix and Mixup \cite{guo2019mixup} methods to enhance model performance and generalization in deep learning tasks. Random erasing or cutout \cite{zhong2020random,devries2017improved} adds noise block to the image to regularize the network, while it helps to solve the occlusion problem. 

Please note that our approach differs fundamentally in motivation and effect from the methods mentioned above. For instance, Random erasing primarily aims to address the issue of decreased generalization performance when the target is occluded. Although this method appears to eliminate color, it actually directly disrupts the structural information of the image. Therefore, the model cannot learn color invariance from relevant parts with or without color. In fact, when the camera style changes, it may even degrade the model's original performance. This observation is further validated in our cross-domain experiments.

With the increasing maturity of GANs, GANs-based approaches for data augmentation have become an active research field.
\subsection{GAN-Based Methods}	
In ReID, the goal of GANs methods \cite{gu2022clothes,zheng2019joint,Camera-style,liu2018pose} is to mitigate the effect of clothing change of pedestrians, color deviation or human-pose variation, and to improve the robustness of the model by learning the invariant features from the variation of the input. The appearance details and the emphases generated by different GANs-based methods are also different, but their goal is all to compensate for the difference between the source and target domains. For example, CamStyle \cite{Camera-style} generates new data for transferring different camera styles to learn invariant features between different cameras to increase the robustness of the model to camera style changes. CycleGAN \cite{zhu2017unpaired} was applied in \cite{deng2018image,zhong2019invariance} to transfer pedestrian image styles from one dataset to another. StarGAN \cite{choi2018stargan} was used by \cite{zhong2018generalizing} to generate pedestrian images with different camera styles. Wei et al. \cite{wei2018person} proposed PTGAN to achieve pedestrian image transfer across different ReID datasets. It uses semantic segmentation to extract foreground masks to assist style transfer, and converts the background into the desired style of the dataset while keeping the foreground unchanged. Different from global style transfer, DGNet \cite{zheng2019joint} utilized GANs to transfer clothing color among different pedestrians to generate more diverse data to reduce the impact of color changes on the model, which effectively improves the generalization ability of the model. CCFA~\cite{han2023clothing} is a variant of DGNet. In contrast to DGNet, which primarily emphasizes learning color invariance, CCFA places greater emphasis on addressing the challenge of insufficient diversity in clothing styles within the training data. Therefore, in the experiments conducted in this paper, our primary focus is to compare with DGNet.

Up to now, GAN-based methods in ReID are considered the optimal choice for enhancing the model's robustness to color variations. By exploring color invariance in ensemble learning, we now present a novel solution in our research.

\begin{algorithm}[h]
	\caption{Random Color Erasing}
	\label{alg:randomcolorerasing}
	\textbf{Input}: Input image $x_i^{v}$; Global erasing probability $p_g$; Erasing probability $p_r$.
	
	\textbf{Output}: Color erased image $I$.
	
	\begin{algorithmic}[1]
		\STATE \textbf{Initialize} $p \leftarrow $ Rand (0, 1).
		\IF{$p \geq p_r$}
		\RETURN $x_i^{v}$.
		\ELSE
		\STATE $p \leftarrow $ Rand (0, 1).
		\IF{$ p \leq p_g$}
		\RETURN $I\leftarrow t(x_i^{v})$.
		\ELSE

		\STATE $x_i^{g} \leftarrow t(x_i^{v})$.
		\STATE $rect = RandPosition(x_i^{v})$.
		\STATE $I \leftarrow $ $LT(x_i^{v},x_i^{g},rect)$.
		\RETURN $I$.

		\ENDIF
		\ENDIF
	\end{algorithmic}
\end{algorithm}

\section{Proposed Methods}
\label{sec:}
The proposed strategy RCE includes both global and local color erasing. Global color erasing can be viewed as a special case of random color erasing. The corresponding analysis of the proposed method is provided at the end of this subsection. The diagram illustrating the proposed method is presented in Fig.~\ref{LGT-LST}. The procedure of RCE is outlined in Alg.1.

\subsection{Global Color Erasing}

In the data loading, it randomly samples M images of per person and K identities to constitute a training batch, which size is $B=M\times K$. The set is denoted as $x^{v}=\{x_i^{v}|i=1,2,...,M\times K\}$.

The proposed method randomly performs global grayscale transformation on the training batch with a probability, and then inputs the processed images into the model for training. This grayscale transformation process can be defined as:
\begin{equation}
	x_i^{g} = t(x_i^{v})
\end{equation}
where $t(\bullet)$ is the grayscale image conversion function, which is implemented pixel-by-pixel accumulation calculations on the R, G, and B channels of the original visible RGB image. 

\subsection{Local Color Erasing}

\begin{figure*}[htbp]
	\setlength{\abovecaptionskip}{0.1cm}
	\setlength{\belowcaptionskip}{-0.4cm}   
	\centering
	\includegraphics[width=0.8\linewidth]{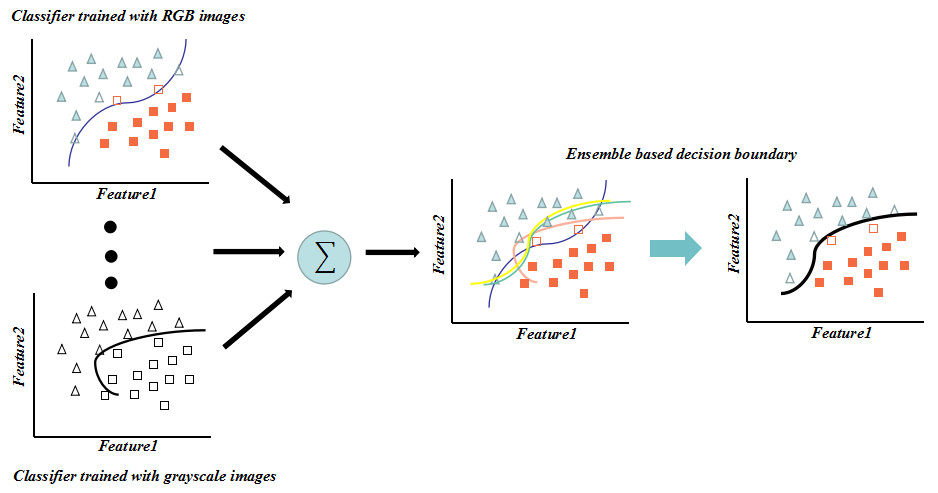}
	\caption{Analysis of Random Color Erasing Strategy. Assuming that the prediction of component networks are combined by majority voting, each component network is regard as a classifier. The component predictions are combined via weighted voting for classification, where the weights are determined by the algorithms themselves. Some of classifiers are trained using visible images, while others are trained using grayscale images. }
	\label{analysis}
\end{figure*}

The local color erasing for each visible image $x_i^v$ can be achieved by the following equations:

\begin{equation}
	rect = RandPosition(x_i^{v}),
\end{equation}

\begin{equation}
	x_i^{lg} = LT(x_i^{v},x_i^{g},rect).
\end{equation}
the function $LT(\bullet)$ can be represented as:
\begin{equation}
	LT(x^{v}_i,x^{g}_i,rect) =  x^{v}_i - x^{v}_i(rect) + x^{g}_i(rect).
\end{equation}
$RandPosition(\bullet)$ is used to generate a random rectangle in the image, the function of $LT(\bullet)$ is to replace the pixels at the rectangular position in the image $x_i^g$ with the pixels at the corresponding position in the  image $x_i^v$. And $x_i^{lg}$ is the sample after local grayscale transformation.

In the process of model training, we apply a local grayscale transformation function $LT(\bullet)$, randomly to the training batch with a certain probability. For an image $x_i^{v}$ within a batch, $p_r$ represents the probability of performing $LT(\bullet)$ on the image $x_i^{v}$. During this process, a rectangular region in the image is randomly selected and replaced with pixels from the corresponding rectangular region in the grayscale image. This introduces various levels of grayscale information into the training images. This process generates training images with diverse levels of grayscale while preserving the structural integrity of the objects. For specific details on the random selection of rectangular regions, please refer to random erasing~\cite{zhong2020random}. We adopt a similar process and parameter settings.


\subsection{Analysis of Random Color Erasing Strategy}
Assuming this task is to approximate a function $f: Q^m \rightarrow Y$ using an ensemble composed of $N$ component neural networks, where $Y$ is the set of class labels, and the prediction of component networks are combined by majority voting, in which each component network votes for a class, and the class label with the largest number of votes is obtained as the output of the ensemble. For the convenience of discussion, assuming that $Y$ contains only two class labels. The set of two class labels is usually denoted as \{0,1\}, but for the convenience of derivation, here use \{-1, +1\} instead of \{0,1\} to represent the class label. So the function to be approximated is $f: Q^m \rightarrow \{-1,   +1\}$. Nevertheless, please note that the following derivation can also be extended to the case where $Y$ contains more than two class labels.

Assuming there are $k$ samples, the expected output $O = [o_1, o_2, …, o_k]^T$, where $d_j$ denotes the expected output on the $j$-th instance, and the actual output of the $i$-th component neural network is $F_i=[f_{i1}, f_{i2}, …, f_{ik}]^T$ where $f_{ij}$ denotes the actual output of the $i$-th component network on the $j$-th sample. $O$ and $F_i$ satisfy that $o_j$ $\in \{-1, +1\} (j = 1, 2, …, m)$ and $f_{ij}$$\in\{-1, +1 \} (i = 1, 2, …, N; j = 1, 2, …, k )$, respectively. If the actual output of the $i$-th component network on the $j$-th sample is correct according to the expected output then $f_{ij}o_j = +1$, otherwise $f_{ij}o_j = -1$. Thus the generalization error of the $i$-th component neural network on those $k$ sample is:

\begin{equation}
	E_i = \frac{1}{k}\sum_{j=1}^k {Error(f_{ij}o_j)}
\end{equation}
where $Error(\bullet)$ is a function defined as:
\begin{equation}
	Error(x)=\left\{
	\begin{array}{ll}
		1, \quad\quad if \quad x=-1\\
		0, \quad\quad if \quad x=1
	\end{array}
	\right.
\end{equation}

Here we introduce a vector $S = [S_1, S_2, …, S_k]^T$ where $S_j$ denotes the sum of the actual output of all the component neural networks on the $j$-th sample:

\begin{equation}
	S_j = \sum_{i=1}^N f_{ij}
\end{equation}

Then the output of the neural network ensemble on the j-th sample is:

\begin{equation}
	\hat{f_j} = Sgn(S_j)
\end{equation}

where $\hat{f}_j\in \{-1, 0, +1 \} (j = 1, 2, …, k)$, and $Sgn(\bullet)$ is a function defined as:

\begin{equation}
	Sgn(x)=\left\{
	\begin{array}{ll}
		1, \quad\quad if \quad x>0\\
		0, \quad\quad if \quad x=0\\
		-1, \quad if \quad x<0
	\end{array}
	\right.
\end{equation}

If the output of the ensemble on the $j$-th sample is correct according to the expected output then $ \hat{f_j}o_j = +1$. If it is wrong then $ \hat{f}_jo_j = -1$. Otherwise, $ \hat{f}_jo_j = 0$, which means that there is a tie on the j-th sample. It means that half component networks vote for +1 while the other half networks vote for -1. Thus the generalization error of the ensemble is:

\begin{equation}
	\label{e-18}
	\hat{E} = \frac{1}{k}\sum_{j=1}^k {Error(\hat{f_{j}}o_j)}
\end{equation}

Here suppose that the $e$-th component neural network is trained using grayscale images. Then the output of the new ensemble on the $j$-th instance is:
\begin{equation}
	\hat{f_j^{'}} = Sgn(S_{j(j \neq e)} + f_{ej})
\end{equation}
and the generalization error of the new ensemble is:
\begin{equation}
	\label{e-20}
	\hat{E^{'}} = \frac{1}{k}\sum_{j=1}^k {Error(\hat{f_{j}^{'}}o_j)}
\end{equation}
Assuming a certain quantity of networks with color information deviation will not impact the overall performance of the neural network, and neglecting color information may lead to improved retrieval results for certain examples.
\begin{equation}
	\label{e-21}
	Error(\hat{f_j^{'}}o_j) \leqslant Error(\hat{f_j}o_j)
\end{equation}
From Eq.~(\ref{e-18}) and Eqn.~(\ref{e-20}) we can derive that if Eqn.~(\ref{e-21}) is satisfied then $\hat{E}$ is not smaller than $\hat{E^{'}}$, and then:
\begin{equation}
	\label{e-22}
	\begin{split}
		\sum_{j=1}^k\{Error(Sgn(S_j)o_j)- \\Error(Sgn(S_{j(j \neq e)}+f_{ej})d_j) \} \geq 0
	\end{split}
\end{equation}
Eqn.~(\ref{e-22}) indicates that the ensemble including $e$-th component neural network which is trained using grayscale images is better than the one no including, as shown in Fig.~\ref{analysis}.

Now, it can be reach the conclusion that in the context of classification, when a number of neural networks are available, ensembling part of them to fit the color features may be better than ensembling all of them. 
\section{Experimental Comparison and Analysis}
In this section, we conducted a sequence of experiments involving two distinct visual tasks, employing five baseline models with varying architectures, and evaluating their performance across five diverse datasets. Additionally, we assessed the effectiveness of six different data augmentation strategies.

\subsection{Person Re-identification Datasets and Evaluation Criteria}
We evaluate our method on three ReID datasets, including Market-1501 \cite{market1501}, DukeMTMC \cite{duke}, MSMT17 \cite{wei2018person}. These three datasets are among the most representative and extensively utilized in ReID research. Together, they encompass multiple seasons, time periods, high-definition, and low-definition cameras, providing rich scenes, backgrounds, and intricate lighting variations.

Following existing works~\cite{market1501}, we employ Rank-k precision and Cumulative Matching Characteristics (CMC) and mean Average Precision (mAP) as evaluation metrics. Rank-1 represents the average accuracy of the top-ranked result corresponding to each cross-modality query image. mAP represents the mean average accuracy, where the query results are sorted based on similarity, and the closer the correct result is to the top of the list, the higher the precision. As part of a ReID system, re-ranking~\cite{re_ranking} technology is typically employed to reorganize the initial retrieval results, aiming to more accurately reflect the similarity between images.

\begin{figure}[h]
	\setlength{\belowcaptionskip}{-0.4cm}   
	\centering
	\includegraphics[width=1\linewidth]{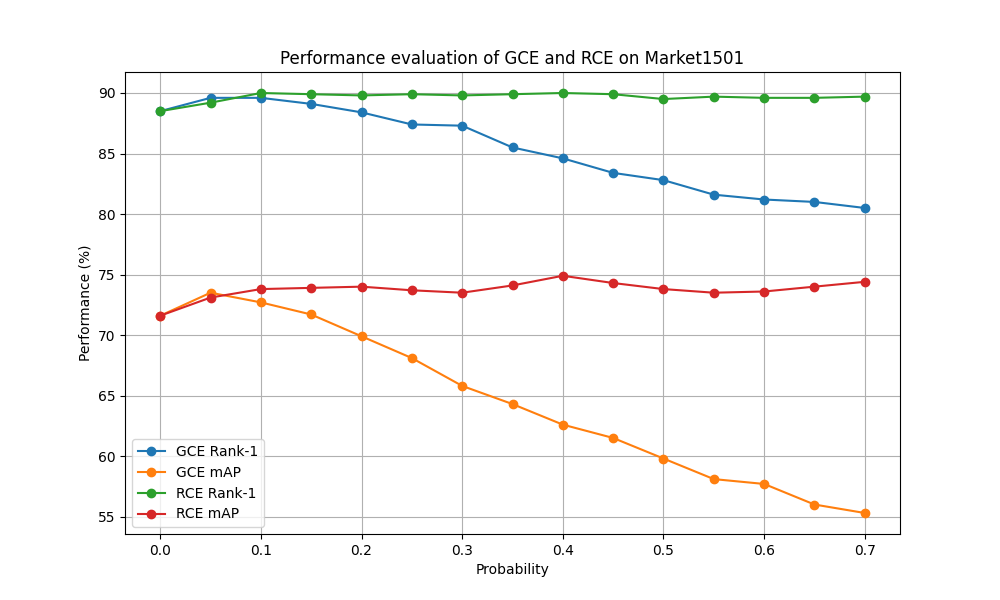}
	\caption{The performance evaluation of our Global Color Erasing (GCE) and our Random Color Erasing (RCE) involves varying hyperparameters on the Market1501 dataset of ReID . Our evaluation results are presented with the default usage of re-ranking.}
	\label{base}
\end{figure}

\subsection{Semantic Segmentation Datasets and Evaluation Criteria}
DustProj consists of 1343 training images, 199 validation images, and 261 testing images. It was annotated by non-experts for training and validation, while professional annotators labeled the test set. The dataset is tailored for practical dust analysis in industrial settings, and it has limitations due to challenging conditions.

The DSS Smoke Segmentation dataset \cite{YUAN2019248} contains 73632 images, with 70632 for training and 3000 for testing, categorized into DS01, DS02, and DS03 groups.

For evaluation, we use mean Intersection over Union (mIoU) as a comprehensive segmentation accuracy metric.

\subsection{Hyper-Parameter Setting and Ablation Study}
During CNN training, two hyper-parameters need to be evaluated. One is global color erasing (GCE) probability $p_g$. The results of different $p_g$ are shown in Fig.~\ref{base}. We can see that when the value $p_g$ is set $0.00$ to $0.15$, the performance of the model is better than the baseline in Rank-1 and mAP. We would like to supplement that when $p_g=0$, it represents the accuracy of the original baseline. When $p_g=1$, it indicates that the model is exclusively trained with grayscale images. The result of $p_g=1$ closely approximates $p_g=0.7$. It indicates that training the model only with grayscale images does not yield satisfactory results.

GCE can be viewed as a specific case of random color erasing (RCE).  Based on the experimental observations mentioned above, we implement GCE with a probability of $p_g=0.15$ within the framework of RCE. When evaluating tatal RCE probability $p_r$, we fixed this parameter. The results of different $p_r$ are shown in Fig.~\ref{base}. Obviously, when $p_r=0.40$, the model achieves the best performance.

\begin{table}[]\small
	\centering 
	\setlength\tabcolsep{2pt}
	\caption{Performance comparison with state-of-the-art methods on the Market1501 and DukeMTMC datasets.}
	\begin{tabular}{ccccc}
		\hline
		\multirow{2}{*}{Methods}   &\multirow{2}{*}{References}       & \multicolumn{1}{c}{Market1501}    & \multicolumn{1}{c}{DukeMTMC}         \\  \cline{3-4} 
		&           & Rank-1/mAP(\%)      	& Rank-1/mAP(\%)            \\  \hline 
		HOReID         &CVPR(2020)     & 94.2/84.9     & 86.9/75.6               \\ 
		OfM       &AAAI(2021)          & 94.9/87.9         & 89.0/78.6            \\ 
		PAT        &CVPR(2021)          & 95.4/88.0     & 88.8/78.2 \\ 
		DRL-Net    &TMM(2022)  &94.7/86.9     & 88.1/76.6   \\ 
		DCAL    &CVPR(2022)   & 94.7/87.5         & 89.0/80.1        \\ 
		CLIP-ReID  &AAAI(2023)   & 95.4/90.5      & 90.8/83.1             \\ \hline
		FastReID      &ACM MM(2023)       & 96. 3/90.3         & 92.4/83.2        \\ 
		FastReID+RCE  &(ours)    & \textbf{96.5/91.2} & \textbf{92.8/84.2}  \\ \hline
	\end{tabular}
	\label{sota1}
\end{table}

\begin{table}[]\small
	\centering 
	\setlength\tabcolsep{2pt}
	\caption{Performance comparison with different methods on the MSMT17 dataset.}
	\begin{tabular}{ccc}
		\hline
		\multirow{2}{*}{Methods}      & \multicolumn{2}{c}{MSMT17}                \\ \cline{2-3} 
		& Rank-1(\%)          & mAP(\%)             \\ \hline
		FastReID           & 82.1                & 60.6             \\ 
		FastReID+REA           & 85.1               & 63.8             \\ 
		FastReID+REA+AutoAugment           & 84.2                & 61.4             \\ 
		FastReID+REA+RCE(ours)     & \textbf{86.2} & \textbf{65.9} \\ \hline
	\end{tabular}
	\label{t3}
\end{table}

\begin{table}[]\small
	\centering 
	\setlength\tabcolsep{2pt}
	\caption{Performance comparison on the Market1501 dataset. +RK indicates the use of re-ranking.}
	\begin{tabular}{ccc}
		\hline
		\multirow{2}{*}{Methods}         & \multicolumn{1}{c}{Market1501}    & \multicolumn{1}{c}{DukeMTMC}         \\  \cline{2-3} 
		& Rank-1/mAP(\%)      	& Rank-1/mAP(\%)            \\  \hline 
		SB            & 94.5/85.9     & 86.4/ 76.4               \\ 
		SB+RK                & 95.4/94.2         & 90.3/89.1            \\ 
		SB+RCE(ours)             & \textbf{95.1/87.2}     & \textbf{87.8/77.3} \\ 
		SB+RCE + RK(ours)    & \textbf{95.9/94.4}    & \textbf{91.0/89.4} \\ \hline
		FastReID       & 96. 3/90.3         & 92.4/83.2        \\ 
		FastReID+RK     & 96.8/95.3      & 94.4/92.2             \\ 
		FastReID+RCE(ours)             & \textbf{96.5/91.2}    & \textbf{92.8/84.2} \\ 
		FastReID+RCE+RK(ours)    & \textbf{96.9/95.6} & \textbf{94.3/92.7}  \\ \hline
	\end{tabular}
	\label{t1}
\end{table}

\begin{table}[]\small
	\centering 
	\setlength\tabcolsep{12pt}
	\caption{Performance comparison between our RCE and DGNet on Market1501.}
	\begin{tabular}{cclcl}
		\hline
		\multirow{2}{*}{Methods} & \multicolumn{4}{c}{Market1501}                                        \\ \cline{2-5} 
		& \multicolumn{2}{c}{Rank-1}        & \multicolumn{2}{c}{mAP(\%)}       \\ \hline
		baseline            & \multicolumn{2}{c}{88.5}          & \multicolumn{2}{c}{71.6}          \\ \hline
		baseline+DGNet       & \multicolumn{2}{c}{{88.9}} & \multicolumn{2}{c}{{72.1}} \\ 
		baseline+RCE(ours) & \multicolumn{2}{c}{\textbf{90.0}} & \multicolumn{2}{c}{\textbf{74.9}} \\\hline
	\end{tabular}
	\label{t5}
\end{table}

The results of the ablation experiments are also reflected in Fig.~\ref{base}. It shows that our GCE outperforms the baseline within a probability range of 0.15. Combining local and global color erasing, referred to as RCE, further enhances the baseline. With RCE, we achieve a 1.5 percentage point improvement in Rank-1 and 3.3 percentage point improvement in mAP over the baseline, and a 2.0 percentage point improvement in Rank-1 and 2.7 percentage point improvement in mAP.

\subsection{Comparison Experiments on Person Re-identification}
We extend our evaluation to State-of-the-Art ReID baselines. As demonstrated in Tab.~\ref{t1}, our method exhibits improvements of 0.6 and 1.3 percentage points in Rank-1 accuracy and mean average precision (mAP) for SB \cite{stong_baseline} on Market1501. Moreover, we observe enhancements of 0.2 percentage points in Rank-1 accuracy and 0.9 percentage points in mAP for FastReID \cite{he2020fastreid}. Importantly, these improvements hold consistently across both DukeMTMC and MSMT17 datasets (see Tab.~\ref{t3}).

\begin{table}[]\small
	\centering 
	\setlength\tabcolsep{2pt}
	\caption{Cross-domain tests. M→D means that we train the model on Market1501 and evaluate it on DukeMTMC, D→M means the reverse.}
	\begin{tabular}{cclcl}
		\hline
		\multirow{3}{*}{Methods}       & \multicolumn{4}{c}{Cross-Domain}                                                \\ \cline{2-5} 
		& \multicolumn{2}{c}{M→D}  & \multicolumn{2}{c}{D→M}  \\ \cline{2-5} 
		& \multicolumn{2}{c}{Rank-1/mAP(\%)}     & \multicolumn{2}{c}{Rank-1/mAP(\%)}     \\ \hline
		baseline     & \multicolumn{2}{c}{37.8/27.0}    & \multicolumn{2}{c}{51.2/31.8}     \\ 
		baseline+REA       & \multicolumn{2}{c}{{29.5/18.4}}    & \multicolumn{2}{c}{43.6/24.1}    \\ 
		baseline+DGNet       & \multicolumn{2}{c}{{36.7/25.6}}    & \multicolumn{2}{c}{52.4/31.6}    \\ 
		baseline+RCE(ours) & \multicolumn{2}{c}{\textbf{39.7/27.9}}     & \multicolumn{2}{c}{\textbf{55.1/35.2}}     \\ \hline
		SB           & \multicolumn{2}{c}{45.5/37.0}          & \multicolumn{2}{c}{58.2/37.8}          \\ 
		SB+REA           & \multicolumn{2}{c}{33.6/24.3}          & \multicolumn{2}{c}{51.6/32.3}          \\ 
		SB+REA+RCE(ours) & \multicolumn{2}{c}{\textbf{37.8/27.8}} & \multicolumn{2}{c}{\textbf{55.4/35.7}} \\ 
		SB+RCE(ours) & \multicolumn{2}{c}{\textbf{48.2/37.9}} & \multicolumn{2}{c}{\textbf{65.0/43.7}} \\ \hline
	\end{tabular}
	\label{t4}
\end{table}

\textbf{Comparison with State-of-the-Art methods}. We compare our method with the State-of-the-Art methods, including HOReID \cite{wang2020high}, OfM \cite{zhang2021one}, PAT \cite{li2021diverse}, DRL-Net \cite{jia2022learning}, DCAL \cite{zhu2022dual} and CLIP-ReID \cite{li2023clip}. As shown in Tab.~\ref{sota1}, we achieve state-of-the-art performance on FastReID by using proposed learning strategy RCE.

Our method enhances accuracy beyond default configurations using data augmentation in FastReID, which includes AutoAugment~\cite{47890} and random erasing augmentation (REA)~\cite{zhong2020random}. Notably, the automated data augmentation strategy, AutoAugment, which employs a series of augmentation techniques to simulate color deviations, does not consistently result in improved model performance.

Another comparison of the performance of our method with State-of-the-Art GAN-Based ReID methods is shown in Tab.~\ref{t5} and Tab.~\ref{t4}. As can be seen from Tab.~\ref{t5}, our method delivers a performance improvement that far exceeds that of DGNet, the GAN-based method, by more than 2.8 percentage points on mAP. As can be seen from Tab.~\ref{t4}, the generalization ability of the proposed method in M→D cross-domain tests is improved by 1.9 percentage points in the Rank-1 compared with the baseline. The improvement is even more pronounced in the M→D cross-domain tests, which further shows that the proposed method is better than DGNet. Notably, using data generated by DGNet for model training leads to poor cross-domain performance.

\begin{table}[t]
	\small
	\setlength\tabcolsep{4pt}%
	\centering
	\caption{Performance comparison of two strong baselines on the DustProj and DSS datasets.}
	\begin{tabular}{cllll}
		\hline
		\multirow{2}{*}{Methods} & \multicolumn{1}{c}{MIoU (\%)}  & \multicolumn{3}{c}{MIoU (\%) in DSS} \\ \cline{3-5}
		& \multicolumn{1}{c}{in DustProj}   & \multicolumn{1}{c}{DS01} & \multicolumn{1}{c}{DS02} & \multicolumn{1}{c}{DS03}  \\ \hline 
		{Segformer}                  & \multicolumn{1}{c}{59.2}                   & \multicolumn{1}{c}{74.8} & \multicolumn{1}{c}{75.4} & \multicolumn{1}{c}{75.1} \\
		{DDRNet}                      & \multicolumn{1}{c}{53.0}                  & \multicolumn{1}{c}{75.0} & \multicolumn{1}{c}{74.2} & \multicolumn{1}{c}{73.1} \\ \hline
		{Segformer+RCE(ours)}       & \multicolumn{1}{c}{60.5}                    & \multicolumn{1}{c}{78.2} & \multicolumn{1}{c}{77.7} & \multicolumn{1}{c}{77.5} \\
		{DDRNet+RCE(ours)  }         & \multicolumn{1}{c}{\textbf{60.9}}          & \multicolumn{1}{c}{\textbf{79.7}} & \multicolumn{1}{c}{\textbf{79.6}} & {\textbf{80.7}} \\ \hline
	\end{tabular}
	\label{SS1}
\end{table}

\textbf{Cross-domain tests}. Cross-domain ReID aims to adapt model training from a labeled source domain dataset to a target domain dataset. However, achieving higher accuracy does not always translate to better generalization, as highlighted by Luo \cite{stong_baseline}. To investigate the effectiveness of our method in cross-domain experiments, we compare RCE with other methods for cross-domain tests between Market-1501 and DukeMTMC, as illustrated in Tab.~\ref{t4}.

As shown in Table~\ref{t4}, due to the direct disruption of sample structural information during the model learning process by random erasing augmentation (REA), it results in adverse effects in cross-domain scenarios, consequently undermining the model's performance. In contrast, our RCE demonstrates more pronounced gains in cross-domain testing through ensemble learning at the image level. This clearly indicates that the proposed method effectively guides the model to learn color invariance. This observation emphasizes the exceptional capability of our method in enhancing model robustness against color variations.

\subsection{Visualization Analysis}  
\begin{figure}[h]
	\setlength{\belowcaptionskip}{-0.4cm}   
	\centering
	\includegraphics[width=1\linewidth]{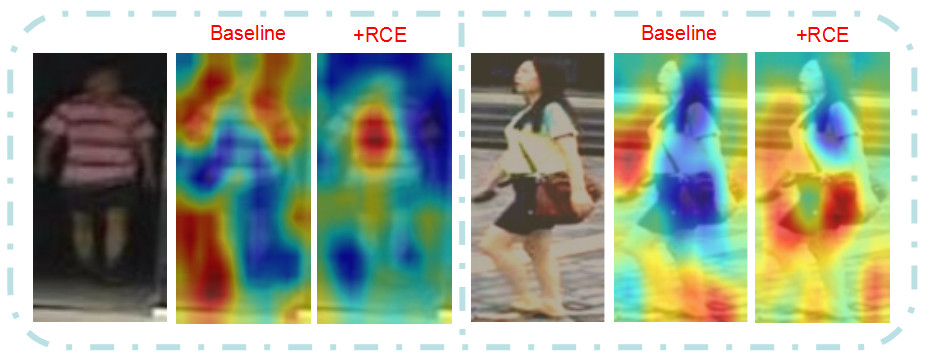}
	\caption{Comparison of Grad-CAM activation maps between a normally trained model and a model trained using our Random Color Erasing.}
	\label{Grad-CAM}
\end{figure}

Grad-CAM \cite{Grad-CAM} employs gradients from the final CNN convolutional layer to visualize neuron importance in the output prediction. It highlights regions influencing model predictions. As shown in Fig.~\ref{Grad-CAM}, the normally trained model activates irrelevant areas under color deviation, whereas RCE-trained model effectively activates critical regions. This indicates that RCE exhibits significant robustness against color deviations.

\begin{table}[t]\small
	\setlength\tabcolsep{8pt}%
	\centering
	\caption{Comparison of different data augmentation methods on the DSS dataset.}
	\begin{tabular}{cccc}
		\hline
		\multirow{2}{*}{Methods}  & \multicolumn{3}{c}{MIoU (\%) in DSS} \\ \cline{2-4}
		& DS01 & DS02 & DS03  \\ \hline 
		DDRNet                & 75.0 & 74.2 & 73.1 \\ 
		DDRNet+Cutout         & 75.5 & 73.8 & 75.1 \\ 
		DDRNet+Co-Mixup         & 77.4 & 76.9 & 77.2 \\  \hline 
		DDRNet+REC(ours)           & \textbf{79.7} & \textbf{79.6} & \textbf{80.7} \\ \hline
	\end{tabular}
	\label{SS2}
\end{table}

\subsection{Comparison Experiments on Semantic Segmentation}
Observing Tab.\ref{SS1}, it can be noted that on both the dust and smoke semantic segmentation datasets, our method consistently demonstrates significant improvements across different architectures of strong baselines, including Segformer~\cite{xie2021segformer} and DDRNet~\cite{pan2022deep}. Notably, our method achieves improvements of over 7 percentage points on DDRNet in both datasets (DustProj and DS03). In addition, we compare our RCE with two methods, namely Cutout and Co-Mixup in Tab.~\ref{SS2}. The experiments demonstrate that our approach achieves superior performance in this task.

Our approach consistently improves the results of various baseline models across different tasks and datasets, demonstrating the effectiveness of the proposed method.


\section{Conclusion}
This paper introduces a novel image-level ensemble learning method, namely Random Color Erasing (RCE), to enhance the robustness of deep learning models in various visual scenarios encountering color variations. By effectively rebalancing the emphasis between color features and other discriminative features, RCE demonstrates its potential in improving model generalization performance. We provide an interpretable analysis of the proposed method, and experiments conducted on ReID and industrial dust (or smoke) segmentation across different architectures and datasets validate the effectiveness of our approach. In conclusion, we aim for this research to serve as a valuable foundation for advancing the field of computer vision.
\bibliographystyle{named}
\bibliography{ijcai24}

\begin{thebibliography}{}

\bibitem[\protect\citeauthoryear{Choi \bgroup \em et al.\egroup
  }{2018}]{choi2018stargan}
Yunjey Choi, Minje Choi, Munyoung Kim, Jung-Woo Ha, Sunghun Kim, and Jaegul
  Choo.
\newblock Stargan: Unified generative adversarial networks for multi-domain
  image-to-image translation.
\newblock In {\em Proceedings of the IEEE conference on computer vision and
  pattern recognition}, pages 8789--8797, 2018.

\bibitem[\protect\citeauthoryear{Cubuk \bgroup \em et al.\egroup
  }{2019}]{47890}
Ekin~Dogus Cubuk, Barret Zoph, Dandelion Mane, Vijay Vasudevan, and Quoc~V. Le.
\newblock Autoaugment: Learning augmentation policies from data.
\newblock In {\em CVPR}, 2019.

\bibitem[\protect\citeauthoryear{Deng \bgroup \em et al.\egroup
  }{2018}]{deng2018image}
Weijian Deng, Liang Zheng, Qixiang Ye, Guoliang Kang, Yi~Yang, and Jianbin
  Jiao.
\newblock Image-image domain adaptation with preserved self-similarity and
  domain-dissimilarity for person re-identification.
\newblock In {\em Proceedings of the IEEE conference on computer vision and
  pattern recognition}, pages 994--1003, 2018.

\bibitem[\protect\citeauthoryear{DeVries and
  Taylor}{2017}]{devries2017improved}
Terrance DeVries and Graham~W Taylor.
\newblock Improved regularization of convolutional neural networks with cutout.
\newblock {\em arXiv preprint arXiv:1708.04552}, 2017.

\bibitem[\protect\citeauthoryear{Gong \bgroup \em et al.\egroup
  }{2022}]{Gong_2022_CVPR}
Yunpeng Gong, Liqing Huang, and Lifei Chen.
\newblock Person re-identification method based on color attack and joint
  defence.
\newblock In {\em Proceedings of the IEEE/CVF Conference on Computer Vision and
  Pattern Recognition (CVPR) Workshops}, pages 4313--4322, June 2022.

\bibitem[\protect\citeauthoryear{Gu \bgroup \em et al.\egroup
  }{2022}]{gu2022clothes}
Xinqian Gu, Hong Chang, Bingpeng Ma, Shutao Bai, Shiguang Shan, and Xilin Chen.
\newblock Clothes-changing person re-identification with rgb modality only.
\newblock In {\em Proceedings of the IEEE/CVF Conference on Computer Vision and
  Pattern Recognition}, pages 1060--1069, 2022.

\bibitem[\protect\citeauthoryear{Guo \bgroup \em et al.\egroup
  }{2019}]{guo2019mixup}
Hongyu Guo, Yongyi Mao, and Richong Zhang.
\newblock Mixup as locally linear out-of-manifold regularization.
\newblock In {\em Proceedings of the AAAI conference on artificial
  intelligence}, volume~33, pages 3714--3722, 2019.

\bibitem[\protect\citeauthoryear{Han \bgroup \em et al.\egroup
  }{2023}]{han2023clothing}
Ke~Han, Shaogang Gong, Yan Huang, Liang Wang, and Tieniu Tan.
\newblock Clothing-change feature augmentation for person re-identification.
\newblock In {\em Proceedings of the IEEE/CVF Conference on Computer Vision and
  Pattern Recognition}, pages 22066--22075, 2023.

\bibitem[\protect\citeauthoryear{He \bgroup \em et al.\egroup
  }{2023}]{he2020fastreid}
Lingxiao He, Xingyu Liao, Wu~Liu, Xinchen Liu, Peng Cheng, and Tao Mei.
\newblock Fastreid: A pytorch toolbox for general instance re-identification.
\newblock In {\em Proceedings of the 31st ACM International Conference on
  Multimedia}, pages 9664--9667, 2023.

\bibitem[\protect\citeauthoryear{Jia \bgroup \em et al.\egroup
  }{2022}]{jia2022learning}
Mengxi Jia, Xinhua Cheng, Shijian Lu, and Jian Zhang.
\newblock Learning disentangled representation implicitly via transformer for
  occluded person re-identification.
\newblock {\em IEEE Transactions on Multimedia}, 25:1294--1305, 2022.

\bibitem[\protect\citeauthoryear{Kim \bgroup \em et al.\egroup
  }{2021}]{kim2021comixup}
JangHyun Kim, Wonho Choo, Hosan Jeong, and Hyun~Oh Song.
\newblock Co-mixup: Saliency guided joint mixup with supermodular diversity.
\newblock In {\em International Conference on Learning Representations}, 2021.

\bibitem[\protect\citeauthoryear{Li \bgroup \em et al.\egroup
  }{2014}]{li2014deepreid}
Wei Li, Rui Zhao, Tong Xiao, and Xiaogang Wang.
\newblock Deepreid: Deep filter pairing neural network for person
  re-identification.
\newblock In {\em Proceedings of the IEEE conference on computer vision and
  pattern recognition}, pages 152--159, 2014.

\bibitem[\protect\citeauthoryear{Li \bgroup \em et al.\egroup
  }{2021}]{li2021diverse}
Yulin Li, Jianfeng He, Tianzhu Zhang, Xiang Liu, Yongdong Zhang, and Feng Wu.
\newblock Diverse part discovery: Occluded person re-identification with
  part-aware transformer.
\newblock In {\em Proceedings of the IEEE/CVF Conference on Computer Vision and
  Pattern Recognition}, pages 2898--2907, 2021.

\bibitem[\protect\citeauthoryear{Li \bgroup \em et al.\egroup
  }{2023}]{li2023clip}
Siyuan Li, Li~Sun, and Qingli Li.
\newblock Clip-reid: Exploiting vision-language model for image
  re-identification without concrete text labels.
\newblock In {\em Proceedings of the AAAI Conference on Artificial
  Intelligence}, volume~37, pages 1405--1413, 2023.

\bibitem[\protect\citeauthoryear{Liao \bgroup \em et al.\egroup
  }{2015}]{liao2015person}
Shengcai Liao, Yang Hu, Xiangyu Zhu, and Stan~Z Li.
\newblock Person re-identification by local maximal occurrence representation
  and metric learning.
\newblock In {\em Proceedings of the IEEE conference on computer vision and
  pattern recognition}, pages 2197--2206, 2015.

\bibitem[\protect\citeauthoryear{Liu \bgroup \em et al.\egroup
  }{2018}]{liu2018pose}
Jinxian Liu, Bingbing Ni, Yichao Yan, Peng Zhou, Shuo Cheng, and Jianguo Hu.
\newblock Pose transferrable person re-identification.
\newblock In {\em Proceedings of the IEEE conference on computer vision and
  pattern recognition}, pages 4099--4108, 2018.

\bibitem[\protect\citeauthoryear{Luo \bgroup \em et al.\egroup
  }{2019}]{stong_baseline}
Hao Luo, Youzhi Gu, Xingyu Liao, Shenqi Lai, and Wei Jiang.
\newblock Bag of tricks and a strong baseline for deep person
  re-identification.
\newblock In {\em Proceedings of the IEEE/CVF conference on computer vision and
  pattern recognition workshops}, pages 0--0, 2019.

\bibitem[\protect\citeauthoryear{Pan \bgroup \em et al.\egroup
  }{2023}]{pan2022deep}
Huihui Pan, Yuanduo Hong, Weichao Sun, and Yisong Jia.
\newblock Deep dual-resolution networks for real-time and accurate semantic
  segmentation of traffic scenes.
\newblock {\em IEEE Transactions on Intelligent Transportation Systems},
  24(3):3448--3460, 2023.

\bibitem[\protect\citeauthoryear{Selvaraju \bgroup \em et al.\egroup
  }{2017}]{Grad-CAM}
Ramprasaath~R. Selvaraju, Michael Cogswell, Abhishek Das, Ramakrishna Vedantam,
  Devi Parikh, and Dhruv Batra.
\newblock Grad-cam: Visual explanations from deep networks via gradient-based
  localization.
\newblock In {\em ICCV}, pages 618--626, 2017.

\bibitem[\protect\citeauthoryear{Wang \bgroup \em et al.\egroup
  }{2020}]{wang2020high}
Guan'an Wang, Shuo Yang, Huanyu Liu, Zhicheng Wang, Yang Yang, Shuliang Wang,
  Gang Yu, Erjin Zhou, and Jian Sun.
\newblock High-order information matters: Learning relation and topology for
  occluded person re-identification.
\newblock In {\em Proceedings of the IEEE/CVF conference on computer vision and
  pattern recognition}, pages 6449--6458, 2020.

\bibitem[\protect\citeauthoryear{Wei \bgroup \em et al.\egroup
  }{2018}]{wei2018person}
Longhui Wei, Shiliang Zhang, Wen Gao, and Qi~Tian.
\newblock Person transfer gan to bridge domain gap for person
  re-identification.
\newblock In {\em Proceedings of the IEEE conference on computer vision and
  pattern recognition}, pages 79--88, 2018.

\bibitem[\protect\citeauthoryear{Xie \bgroup \em et al.\egroup
  }{2021}]{xie2021segformer}
Enze Xie, Wenhai Wang, Zhiding Yu, Anima Anandkumar, Jose~M Alvarez, and Ping
  Luo.
\newblock Segformer: Simple and efficient design for semantic segmentation with
  transformers.
\newblock {\em Advances in Neural Information Processing Systems},
  34:12077--12090, 2021.

\bibitem[\protect\citeauthoryear{Yuan \bgroup \em et al.\egroup
  }{2019}]{YUAN2019248}
Feiniu Yuan, Lin Zhang, Xue Xia, Boyang Wan, Qinghua Huang, and Xuelong Li.
\newblock Deep smoke segmentation.
\newblock {\em Neurocomputing}, 357:248--260, 2019.

\bibitem[\protect\citeauthoryear{Yuan \bgroup \em et al.\egroup
  }{2023}]{yuan2023lightweight}
Feiniu Yuan, Kang Li, Chunmei Wang, and Zhijun Fang.
\newblock A lightweight network for smoke semantic segmentation.
\newblock {\em Pattern Recognition}, 137:109289, 2023.

\bibitem[\protect\citeauthoryear{Yun \bgroup \em et al.\egroup
  }{2019}]{yun2019cutmix}
Sangdoo Yun, Dongyoon Han, Seong~Joon Oh, Sanghyuk Chun, Junsuk Choe, and
  Youngjoon Yoo.
\newblock Cutmix: Regularization strategy to train strong classifiers with
  localizable features.
\newblock In {\em Proceedings of the IEEE/CVF international conference on
  computer vision}, pages 6023--6032, 2019.

\bibitem[\protect\citeauthoryear{Yunpeng~Gong}{2024}]{gong2024}
Zhiming Luo Yansong Qu Rongrong Ji Min~Jiang Yunpeng~Gong, Zhun~Zhong.
\newblock Cross-modality perturbation synergy attack for person
  re-identification.
\newblock {\em https://arxiv.org/pdf/2401.10090.pdf}, 2024.

\bibitem[\protect\citeauthoryear{Zhang \bgroup \em et al.\egroup
  }{2021}]{zhang2021one}
Enwei Zhang, Xinyang Jiang, Hao Cheng, Ancong Wu, Fufu Yu, Ke~Li, Xiaowei Guo,
  Feng Zheng, Weishi Zheng, and Xing Sun.
\newblock One for more: Selecting generalizable samples for generalizable reid
  model.
\newblock In {\em Proceedings of the AAAI Conference on Artificial
  Intelligence}, volume~35, pages 3324--3332, 2021.

\bibitem[\protect\citeauthoryear{Zheng \bgroup \em et al.\egroup
  }{2015}]{market1501}
Liang Zheng, Liyue Shen, Lu~Tian, Shengjin Wang, Jingdong Wang, and Qi~Tian.
\newblock Scalable person re-identification: A benchmark.
\newblock In {\em Proceedings of the IEEE international conference on computer
  vision}, pages 1116--1124, 2015.

\bibitem[\protect\citeauthoryear{Zheng \bgroup \em et al.\egroup }{2017}]{duke}
Zhedong Zheng, Liang Zheng, and Yi~Yang.
\newblock Unlabeled samples generated by gan improve the person
  re-identification baseline in vitro.
\newblock In {\em Proceedings of the IEEE international conference on computer
  vision}, pages 3754--3762, 2017.

\bibitem[\protect\citeauthoryear{Zheng \bgroup \em et al.\egroup
  }{2019}]{zheng2019joint}
Zhedong Zheng, Xiaodong Yang, Zhiding Yu, Liang Zheng, Yi~Yang, and Jan Kautz.
\newblock Joint discriminative and generative learning for person
  re-identification.
\newblock In {\em proceedings of the IEEE/CVF conference on computer vision and
  pattern recognition}, pages 2138--2147, 2019.

\bibitem[\protect\citeauthoryear{Zhong \bgroup \em et al.\egroup
  }{2017}]{re_ranking}
Zhun Zhong, Liang Zheng, Donglin Cao, and Shaozi Li.
\newblock Re-ranking person re-identification with k-reciprocal encoding.
\newblock In {\em CVPR}, 2017.

\bibitem[\protect\citeauthoryear{Zhong \bgroup \em et al.\egroup
  }{2018a}]{zhong2018generalizing}
Zhun Zhong, Liang Zheng, Shaozi Li, and Yi~Yang.
\newblock Generalizing a person retrieval model hetero-and homogeneously.
\newblock In {\em Proceedings of the European conference on computer vision
  (ECCV)}, pages 172--188, 2018.

\bibitem[\protect\citeauthoryear{Zhong \bgroup \em et al.\egroup
  }{2018b}]{Camera-style}
Zhun Zhong, Liang Zheng, Zhedong Zheng, Shaozi Li, and Yi~Yang.
\newblock Camera style adaptation for person re-identification.
\newblock In {\em CVPR}, 2018.

\bibitem[\protect\citeauthoryear{Zhong \bgroup \em et al.\egroup
  }{2019}]{zhong2019invariance}
Zhun Zhong, Liang Zheng, Zhiming Luo, Shaozi Li, and Yi~Yang.
\newblock Invariance matters: Exemplar memory for domain adaptive person
  re-identification.
\newblock In {\em Proceedings of the IEEE/CVF Conference on Computer Vision and
  Pattern Recognition}, pages 598--607, 2019.

\bibitem[\protect\citeauthoryear{Zhong \bgroup \em et al.\egroup
  }{2020}]{zhong2020random}
Zhun Zhong, Liang Zheng, Guoliang Kang, Shaozi Li, and Yi~Yang.
\newblock Random erasing data augmentation.
\newblock In {\em Proceedings of the AAAI conference on artificial
  intelligence}, volume~34, pages 13001--13008, 2020.

\bibitem[\protect\citeauthoryear{Zhou \bgroup \em et al.\egroup
  }{2023}]{zhou2023adaptive}
Xiao Zhou, Yujie Zhong, Zhen Cheng, Fan Liang, and Lin Ma.
\newblock Adaptive sparse pairwise loss for object re-identification.
\newblock In {\em Proceedings of the IEEE/CVF Conference on Computer Vision and
  Pattern Recognition}, pages 19691--19701, 2023.

\bibitem[\protect\citeauthoryear{Zhu \bgroup \em et al.\egroup
  }{2017}]{zhu2017unpaired}
Jun-Yan Zhu, Taesung Park, Phillip Isola, and Alexei~A Efros.
\newblock Unpaired image-to-image translation using cycle-consistent
  adversarial networks.
\newblock In {\em Proceedings of the IEEE international conference on computer
  vision}, pages 2223--2232, 2017.

\bibitem[\protect\citeauthoryear{Zhu \bgroup \em et al.\egroup
  }{2022}]{zhu2022dual}
Haowei Zhu, Wenjing Ke, Dong Li, Ji~Liu, Lu~Tian, and Yi~Shan.
\newblock Dual cross-attention learning for fine-grained visual categorization
  and object re-identification.
\newblock In {\em Proceedings of the IEEE/CVF Conference on Computer Vision and
  Pattern Recognition}, pages 4692--4702, 2022.

\end{thebibliography}

\end{document}